\pdfoutput=1

\documentclass[11pt]{article}

\usepackage[preprint]{acl}

\usepackage{times}
\usepackage{latexsym}
\usepackage{booktabs}
\usepackage{amssymb}
\usepackage{amsmath}
\usepackage{fixltx2e}
\usepackage[most]{tcolorbox} 
\usepackage{listings} 
\usepackage{xcolor} 
\usepackage{listings}
\usepackage{mdframed} 
\definecolor{verylightgray}{gray}{0.97}

\newtcolorbox{promptbox}[1][]{
  colback=verylightgray, 
  colframe=lightgray, 
  title=Optimized Prompt, 
  fonttitle=\small\bfseries, 
  #1,
}

\definecolor{backgroundcolor}{rgb}{0.97, 0.97, 0.97}
\definecolor{delim}{RGB}{20,105,176}
\colorlet{punct}{red!60!black}
\definecolor{numb}{rgb}{0.5,0,0.5}
\definecolor{darkgray}{rgb}{0.4, 0.4, 0.4}
\definecolor{keywordcolor}{rgb}{0.88, 0.4, 0.4}

\lstdefinelanguage{json}{
    basicstyle=\normalfont\ttfamily,
    numberstyle=\scriptsize,
    showstringspaces=false,
    breaklines=true,
    frame=single,
    rulecolor=\color{darkgray}, 
    backgroundcolor=\color{backgroundcolor},
    tabsize=2,
    literate=
     *{0}{{{\color{numb}0}}}{1}
      {1}{{{\color{numb}1}}}{1}
      {2}{{{\color{numb}2}}}{1}
      {3}{{{\color{numb}3}}}{1}
      {4}{{{\color{numb}4}}}{1}
      {5}{{{\color{numb}5}}}{1}
      {6}{{{\color{numb}6}}}{1}
      {7}{{{\color{numb}7}}}{1}
      {8}{{{\color{numb}8}}}{1}
      {9}{{{\color{numb}9}}}{1}
      {:}{{{\color{punct}{:}}}}{1}
      {,}{{{\color{punct}{,}}}}{1}
      {\{}{{{\color{delim}{\{}}}}{1}
      {\}}{{{\color{delim}{\}}}}}{1}
      {[}{{{\color{delim}{[}}}}{1}
      {]}{{{\color{delim}{]}}}}{1},
    morekeywords={role, content, messages},
    keywordstyle=\color{keywordcolor}\bfseries
}

\usepackage[T1]{fontenc}

\usepackage[utf8]{inputenc}
\usepackage{graphicx}

\usepackage{microtype}
\usepackage{algorithm}
\usepackage{algpseudocode}
\usepackage{inconsolata}

\usepackage{graphicx}

\title{ProbGate at EHRSQL 2024: Enhancing SQL Query Generation Accuracy through Probabilistic Threshold Filtering and Error Handling}

\author{Sangryul Kim$^{1\dagger}$~~~ Donghee Han$^{2}$~~~ Sehyun Kim$^{3}$ \smallskip \\
        \textsuperscript{1}KAIST AI\\  \textsuperscript{2}KAIST Graduate School of Data Science\\  \textsuperscript{3}KAIST Bio and Brain Engineering \\
        \texttt{\{sangryul, handonghee, sehyun\}@kaist.ac.kr}\\}

\begin{document}
\maketitle
\begingroup\def\thefootnote{$\dagger$}\footnotetext{Corresponding Author}\endgroup
\begin{abstract}
Recently, deep learning-based language models have significantly enhanced text-to-SQL tasks, with promising applications in retrieving patient records within the medical domain. One notable challenge in such applications is discerning unanswerable queries. Through fine-tuning model, we demonstrate the feasibility of converting medical record inquiries into SQL queries. Additionally, we introduce an entropy-based method to identify and filter out unanswerable results. We further enhance result quality by filtering low-confidence SQL through log probability-based distribution, while grammatical and schema errors are mitigated by executing queries on the actual database.
We experimentally verified that our method can filter unanswerable questions, which can be widely utilized even when the parameters of the model are not accessible, and that it can be effectively utilized in practice\footnote{Code and datasets are available at \url{https://github.com/venzino-han/probgate_ehrsql}}.

\end{abstract}
\section{Introduction}
In recent years, the field of natural language processing (NLP) has witnessed remarkable progress driven by transformer-based large language models (LLMs) \cite{brown2020language, touvron2023llama,roziere2023code}. A prevailing approach involves fine-tuning pre-trained language models with new data across various tasks, facilitating transfer learning \cite{lm_survey}. This methodology has proven effective in tasks like document summarization, entity-relationship extraction, document classification, and sentiment analysis. One of the main tasks where these language models are increasingly leveraged is text-to-SQL (Text2SQL), which converts natural language queries into SQL queries \cite{text2sql_survey_1}.

Text2SQL presents unique challenges distinct from conventional NLP tasks. Firstly, it demands grammatical correctness, as even minor errors can render SQL queries unexecutable. Unlike document summarization, where semantic correctness compensates for grammatical inaccuracies, SQL queries must adhere strictly to syntax rules \cite{syntax_text2sql}. Secondly, schema awareness is crucial; understanding the database structure is essential for generating accurate SQL queries \cite{text2sql_survey_2}. Finally, discerning unanswerable queries is vital, especially in domains like healthcare where incorrect or incomplete information can have severe consequences \cite{lee2022ehrsql}. If users do not inspect the SQL queries themselves, but only receive the results of the execution, the results of an incorrect SQL execution can be fatally misleading.

\begin{figure}[t]
    \centering
    \includegraphics[width=1.0\linewidth]{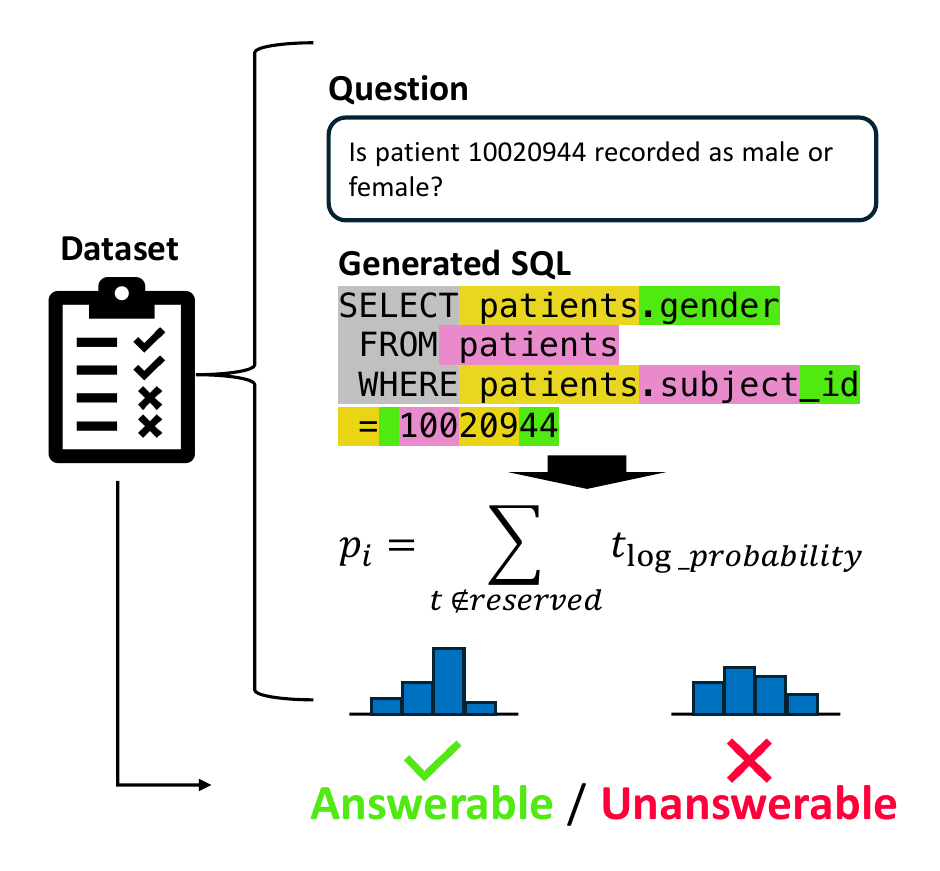}
    \caption{Determines whether a question and the generated SQL are answerable or unanswerable based on the log probability of the tokens generated by the Text2SQL model. If the log probability of a token falls below a certain threshold, we classify the question and SQL as unanswerable. 
    }
    \label{fig:abstract_img}
\end{figure}

In the domain of Text2SQL, effectively filtering out unanswerable questions presents a significant challenge \cite{lee2024trustsql}, particularly within the medical field where accuracy is paramount. Existing methodologies for identifying unanswered queries have primarily targeted cases where such queries exhibit discernible patterns \cite{text2sql_unans}. However, these methods are often tailored to specific model architectures and learning methods, thereby constraining their direct applicability to LLM services accessible via APIs, such as ChatGPT. Given the recent widespread adoption of such managed LLMs across various industries, the need for a more versatile and adaptable approach to filtering unanswered questions becomes increasingly pronounced. This underscores the necessity for innovative solutions that can seamlessly integrate into existing LLM services, ensuring robust performance in diverse application scenarios, including medical contexts.

We address solutions that effectively solve the challenges of Text2SQL tasks through a subset focusing on Electronic Health Records(EHR), utilizing medical questions and corresponding SQL queries relevant to medical systems used in real hospitals \cite{lee2022ehrsql}. Specifically, we participate in the EHRSQL Shared Task on Reliable Text-to-SQL Modeling On Electronic Health Records \cite{lee2024overview}. A distinctive feature of this shared task is that under the basic premise of generating appropriate SQL statements for given natural language queries, not all questions are answerable; some are unanswerable. Moreover, beyond merely generating suitable SQL statements for questions, this task is complex as it requires distinguishing between answerable and unanswerable questions and considering the high penalties for incorrectly identifying the questions are answerable or not, thus necessitating both reliability and accuracy in execution.

In this paper, we introduce \textbf{Prob}ability \textbf{Gate (ProbGate)}, a novel probability-based filtering approach designed for seamless integration with diverse generative language models, without requiring direct access to the model's parameters.
Figure \ref{fig:abstract_img} illustrates the concept of ProbGate, which leverages the logarithmic probability of individual tokens to assess the uncertainty associated with generated SQL queries. We consider the log probability of specific target tokens as an indicator of how confident the model is and how well it can perform the task without hallucinations. We found that utilizing logarithmic probability-based confidence to identify answerable and unanswerable questions was very effective, which is a key aspect of this task.

We evaluate the efficacy of ProbGate through experimentation with Electronic Health Record (EHR) SQL dataset \cite{lee2022ehrsql}. Specifically, we apply ProbGate to both T5-based \cite{2020t5} Text2SQL models and gpt-3.5-turbo finetuned models, comparing their performance against conventional binary classifiers. Additionally, we train binary classifiers based on T5 and gpt-3.5-turbo model  to filter out unaswerable questions. Our experimental findings reveal that ProbGate outperforms binary classifiers in terms of both performance and resilience to shifts in data distribution. These results underscore the potential of ProbGate as a versatile and robust filtering solution for a wide range of applications.

Our contributions and methods can be summarized as follows:
\begin{itemize}
\item{Through our experiments, we found that the fine-tuned gpt-3.5-turbo performed well at generating SQL queries for questions, but was less able to distinguish and filter out unanswerable questions.}
\item{We present the Probabilistic Threshold Filtering method(ProbGate) to effectively distinguish between answerable and unanswerable questions in datasets containing a mix of both. }
\item{We demonstrate an effective method by creating a single pipeline from training to testing, incorporating SQL execution error handling, showing that it can be applied to similar cases.}
\end{itemize}

\section{Backgrounds}
\paragraph{Text2SQL}
Databases serve as powerful tools for efficiently querying extensive datasets. However, accessing this data often requires users to possess knowledge of query languages like SQL. To democratize this process and render it accessible across proficiency levels, significant research efforts have focused on techniques for interpreting natural language questions and autonomously translating them into SQL queries. Recent strides in deep learning methodologies, particularly transformer-based language models, have spurred the development of text-to-SQL techniques. These approaches aim to bridge the gap between natural language queries and SQL commands, thereby enhancing accessibility and usability in database querying tasks \cite{text2sql_survey_1, text2sql_survey_2}.

Early Text2SQL research relied on rule-based and template-based methods, but more recently, deep learning-based methodologies have become mainstream \cite{{deng-etal-2022-recent}}.   
Deep learning methodologies exhibit robustness on the data they are trained on but often struggle to generalize to unseen database schemas. To mitigate this challenge, researchers have explored approaches to encode database relationships and leverage column relationships using self-attention mechanisms \cite{rat-sql}.
In Text2SQL, ensuring the accuracy of generated SQL statements is crucial as even minor errors can lead to failures in query execution. Recent studies have demonstrated the effectiveness of utilizing LLMs like gpt-3.5-turbo to rectify SQL statements derived from natural language queries, addressing the challenge of proofreading SQL output \cite{din_sql}.

One of the main applications of Text2SQL is its utilization in the healthcare domain, specifically to handle complex tasks within electronic health records (EHRs). Recent research has shown that decomposing these tasks into manageable pieces can improve the performance of multi-table reasoning within EHRs. 
The authors proposed to iteratively improve SQL queries by incorporating interactive coding and execution feedback mechanisms to learn from the error messages encountered. This iterative improvement process proved to be effective and resulted in noticeable improvements in SQL performance in the healthcare domain \cite{ehr_agent}.
In a closely related investigation, researchers observed that EHR data is commonly stored in relational databases, which can be represented as directed acyclic graphs. Leveraging this insight, they employed a graph-based methodology to capture the intricate relationships between tables, entities, and values within relational databases \cite{kg_text2sql}.

\paragraph{Confidence of Generated Tokens}
The outputs of LLMs are typically based on a next token prediction method, where the probability of previous tokens is used to predict the next one. During this process, a phenomenon often referred to as `hallucination' can occur, which results in incorrect inferences about the task\cite{wang-sennrich-2020-exposure, xiao-wang-2021-hallucination, li-etal-2022-pre}. Additionaly, previous research has shown that low probability and confidence levels can indicate a lack of knowledge in the model\cite{kadavath2022language}. To overcome this, \citet{jiang-etal-2023-active} introduced a structure named FLARE, which includes a mechanism where if the probability of a token generated by the model falls below a certain threshold, the token is used as a query to retrieve relevant documents from a retriever. This approach aims to address the lack of knowledge and increase confidence. In our work, we also propose a filtering model using log probability to determine if log probability can effectively distinguish the uncertainty in generated content.
\begin{figure*}[t]
    \centering
    \includegraphics[width=1.0\textwidth]{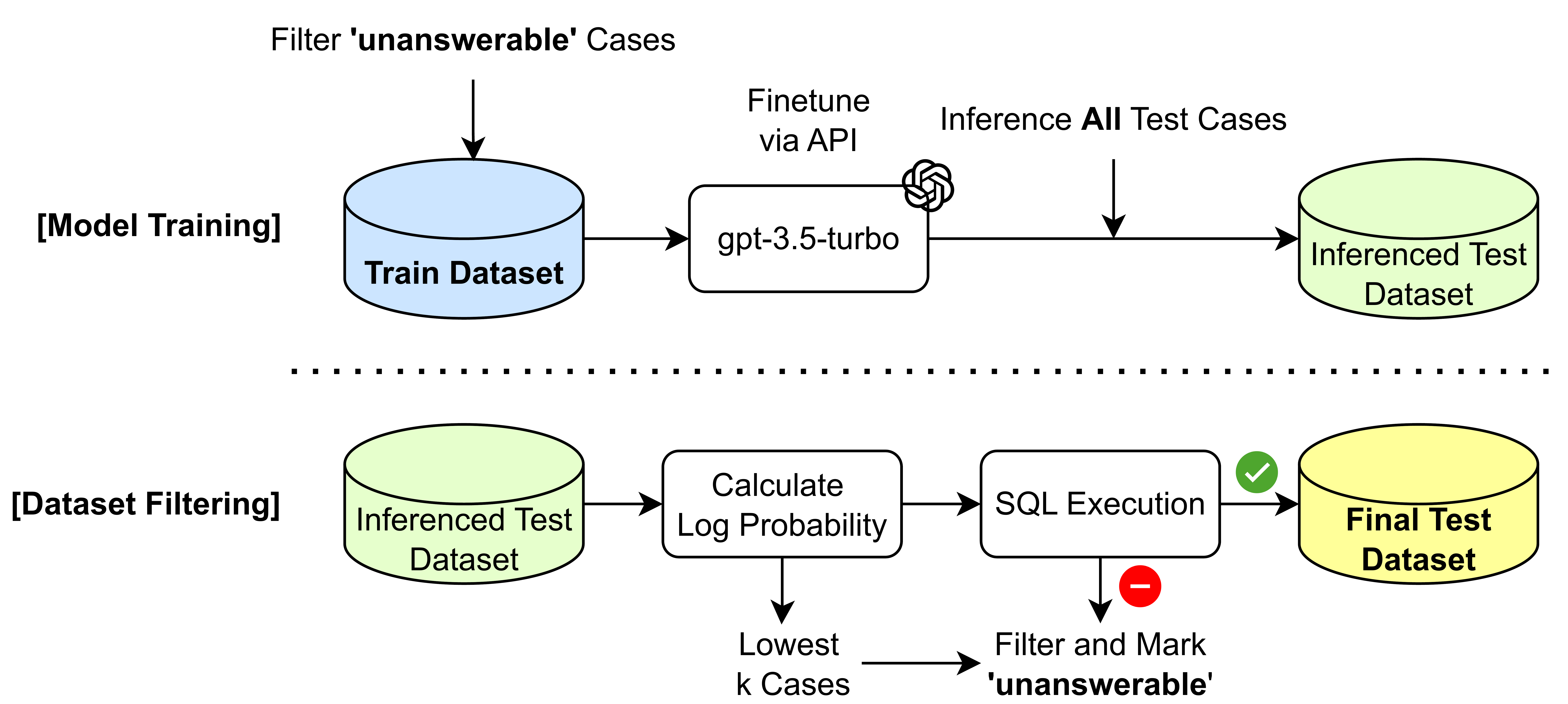}
    \caption{Our method's overall architecture is as follows: During training, we fine-tune the gpt-3.5-turbo model using a dataset from which unanswerable cases have been removed. Subsequently, we identify unanswerable cases using filtering based on log probability and filtering through SQL execution, ultimately deriving the answers.}
    \label{fig:main_img}
\end{figure*}
\section{Methods}

From this section, we cover the contents related to the methods. In \S\ref{sec:methods:datasets}, there is a detailed description of the shared task dataset; in \S\ref{sec:methods:metric}, the main metrics used in the shared task are discussed; and from \S\ref{sec:methods:ftpd} to \S\ref{sec:methods:ehf}, detailed information on the main methods is provided. The entire architecture can be referenced in Figure \ref{fig:main_img}.

\subsection{Datasets}
\label{sec:methods:datasets}
The dataset employed in this study is sourced from the EHRSQL Shared Task on Reliable Text-to-SQL Modeling On Electronic Health Records(EHRSQL-2024) \cite{lee2024overview}, with the purpose of simplifying access to EHR data by automatically translating natural language questions into corresponding SQL queries. This dataset is referred to as The MIMIC-IV demo version of EHRSQL with additional unanswerable questions. It consists of various questions related to medical records and their corresponding SQL queries, serving as a crucial resource for natural language processing and SQL query generation research. The specific attributes and composition follow the study by EHRSQL \cite{lee2022ehrsql}. The EHRSQL dataset is based on questions frequently asked in the medical field, gathered from 222 hospital personnel, including physicians, nurses, insurance assessors, and health records teams. These questions have been reconstructed to reflect various scenarios that can occur in  real-world medical contexts and are presented as a dataset annotated with SQL queries aligned with the hierarchical structure of EHR databases.

 The primary characteristics of this dataset are as follows: it encapsulates the diverse demands of hospital settings, encompassing tasks from straightforward information retrieval to the more intricate operations such as identifying the top N prescribed drugs following a disease diagnosis. Additionally, it incorporates a range of temporal expressions within the questions. Lastly, it includes not only answerable questions but also unanswerable ones that are incompatible with the database schema or require external domain knowledge.
 
 The EHRSQL-2024 task provides a training dataset consisting of questions about medical records, SQL queries corresponding to the MIMIC-IV demo version, and instances annotated as `null' for unanswerable questions. The test dataset comprises only questions, including types of unanswerable questions that are not included in the training data. The training and test datasets comprise 5124 and 1167 examples, respectively.
 
\subsection{Metric}
\label{sec:methods:metric}
In the medical and healthcare domains, reliability is particularly emphasized. Therefore, the model's responses must be accurate, and it's better to abstain from answering than to risk errors. From this perspective, we employ the RS (Reliability Score) introduced in TrustSQL\cite{lee2024trustsql} to assess the model's performance. The RS assigns scores for accurate predictions, providing an evaluation of the model's performance, while also penalizing incorrect predictions and instances where the model attempts to respond to unanswerable questions.

{
\small
\begin{equation}
\phi_c(x) = \begin{cases} 
1 & \text{if } x \in Q_{\text{ans}}; g(x) = 1; \text{Acc}(x) = 1, \\
0 & \text{if } x \in Q_{\text{ans}}; g(x) = 0, \\
-c & \text{if } x \in Q_{\text{ans}}; g(x) = 1; \text{Acc}(x) = 0, \\
-c & \text{if } x \in Q_{\text{una}}; g(x) = 1, \\
1 & \text{if } x \in Q_{\text{una}}; g(x) = 0.
\end{cases}
\label{eq:metric}
\end{equation}
}

In EQ(\ref{eq:metric}), Acc($x$) represents the execution accuracy, where for any $x$ belonging to the set of answerable questions $(Q_{ans})$, if $f(x)$ matches the correct answer, it returns 1, and otherwise, it returns 0. The function $g(x)$ indicates whether the model generates an SQL query, where 1 signifies generation and 0 indicates no generation. The parameter $c$ serves as the penalty parameter. A penalty of $-c$ is imposed in two scenarios: when $x$ is in $Q_{ans}$ and the generated query is incorrect, and when $x$ is in the set of unanswerable questions $(Q_{una})$ but a query is generated regardless. The model earns a score of 1 when it correctly answers a question. The final Reward Score (RS) is obtained by calculating the average of $\phi_{c}(x)$ scores across all samples. The penalty factor $c$ can be adjusted to evaluate the model's reliability, particularly in scenarios requiring high confidence. In our experiments, we consider four options for the penalty, $c = 0, 5, 10, N$, where $N$ represents the total number of samples being evaluated. This metric proves valuable in assessing the model's ability to reliably generate SQL queries and to respond only to questions that are answerable.

\subsection{Fine-Tuning and Prompt Design}
\label{sec:methods:ftpd}
\paragraph{Fine-tuning} As the first step in solving the task, we fine-tune the OpenAI gpt-3.5-turbo-0125 model\footnote{Details of fine-tuning gpt-3.5-turbo model are described at \url{https://platform.openai.com/docs/guides/fine-tuning}}. This is used for Text2SQL conversion, serving as an easy-to-use baseline and also providing a convenient API for subsequent log probability calculations. To minimize noise in the dataset, we exclude unanswerable data from training, focusing solely on SQL transformation without considering whether the given questions are answerable or not. Out of the 5124 samples in the training set, 450 unanswerable data points were excluded, leaving 4674 question-query pairs that are answerable. These data consist of natural language questions paired with their corresponding correct SQL queries. The example of the input-output format for the training dataset can be found in the Appendix \ref{appd:format}.

\paragraph{Prompt} During the training and inference phase, we experiment with various prompt formats to facilitate the model's ability to receive a question and generate the corresponding SQL query accurately. As an illustration, the following structure is utilized for prompts:
\begin{promptbox}
"You are `SQLgpt', an AI designed to convert natural language questions into their corresponding SQL queries. It is imperative that the generated SQL queries conform to the standard SQL format and are not enclosed within quotes (neither single ' nor double "). Your primary objective is to precisely generate the exact SQL query for each presented question."
\end{promptbox}
 Such prompts aim to guide the model towards generating the most appropriate SQL query in response to a question while also preventing the occasional generation of SQL queries encased within ' or " symbols, which can potentially lead to errors within the database.

\subsection{Probabilistic Threshold Filtering (ProbGate)}
\label{sec:methods:probgate}

\begin{algorithm}
\caption{ProbGate}
\label{alg:pth}
\begin{algorithmic}[1]
\State $reserved \gets [``SELECT", \dots]$ 
\Procedure{CalcLogBottomK}{$log$, $t$}
    \State $LogProb \gets []$
    \For{$x$ in $log$}
        \If{$x.token$ not in $reserved$}
            \State $LogProb$.append($x.logprob$)
        \EndIf
    \EndFor
    \State Keep bottom $\textbf{t}$ values of sorted($LogProb$)
    \State \textbf{return} average($LogProb$)
\EndProcedure
\end{algorithmic}
\end{algorithm}

\begin{table*}[h]
\centering
\begin{tabular}{c|cccc}

\toprule
\textbf{Model} & \textbf{RS(0)} & \textbf{RS(5)} & \textbf{RS(10)} &\textbf{Rs(N)}\\
\toprule
T5-small FT + Filtering & 47.81 & 45.66 & 43.51 &\textbf{ -452.19} \\
T5-Large-text2sql-spider FT + Filtering & 74.63 & 59.59 & 44.54 & -3425.37 \\
T5-Large-text2sql-spider FT + Classifier(T5) & 63.80 & 18.23 & -27.34 & -10536.20 \\
T5-Large-text2sql-spider FT + Filtering + Classifier(T5) & 72.74 & 58.56 & 44.37 & -3227.26 \\
gpt-3.5-turbo FT + Classifier(T5) & 90.28 & 51.59 & 12.89 & -8109.02 \\
gpt-3.5-turbo FT + Classifier(gpt-3.5-turbo) & 88.05 & 57.95 & 27.86 & -6911.95 \\
\toprule
gpt-3.5-turbo FT + ProbGate(t=387) & \textbf{85.30}  & \textbf{80.57} & \textbf{75.84} & -1014.70 \\  
\bottomrule
\end{tabular}
\caption{Model Selection and Ablation Study in Dev Phase dataset. In the case of the T5-Large model, it is a model that was first fine-tuned using the Spider dataset, which is one of the Text2SQL datasets, and then subsequently trained on the EHRSQL dataset. In abbreviation, "FT" stands for Fine-Tuning. `Filtering' and `Classifier' are described in section \S\ref{sec:methods:model}.}
\label{tab:ablation}
\end{table*}

\begin{table*}[h]
\centering
{
\begin{tabular}{c|cccc}
\toprule
\textbf{Model} & \textbf{RS(0)} & \textbf{RS(5)} & \textbf{RS(10)} &\textbf{Rs(N)} \\ 
\toprule
gpt-3.5-turbo FT & 73.52 & -58.87 & -191.25 & -30826.47 \\
gpt-3.5-turbo FT + ProbGate(t=450) & 79.43 & 73.01 & 66.58 & -1420.57 \\
gpt-3.5-turbo FT + ProbGate(t=450) + GEF & 79.78 & 75.92 & 72.06 & -820.22 \\
\toprule
\textbf{gpt-3.5-turbo FT + ProbGate(t=425) + GEF} & \textbf{81.92} & \textbf{78.06} & \textbf{74.21} & \textbf{-818.08} \\
\bottomrule
\end{tabular}
}
\caption{The results from applying our methodology during the Test Phase are as follows. The results of ablation at each filtering stage are provided, and it can be observed that there is an improvement in performance at every stage. In abbreviation, "FT" stands for Fine-Tuning, and "GEF" refers to Grammatical Errors Filtering, as introduced in section \S\ref{sec:methods:ehf}. }
\label{tab:final_score}
\end{table*}

In the test set of the given task, we can see that answering all questions as unanswerable results in a score of 19.97 across all RS metrics. By assuming all questions to be answerable and submitting answers accordingly, we were able to achieve a score of 73.52 on the RS(0) metric, in an effort to understand the performance of the model fine-tuned in the previous step on answerable questions. Interpreting this from a ratio perspective, since we already know that 19.97\% of the test set is unanswerable, it implies that 80.03\% of it consists of answerable questions. Therefore, we can deduce that the percent accuracy of the model on answerable questions is approximately 91.87\%. This implies a percent accuracy of 91.87\%, which suggests that to avoid losing points, the threshold for ideally identifying unanswerable questions should be set higher than the scale used to find this threshold, as inferred from the results. Given that the total number of items in the test set for the given task is 1167, we can deduce that to minimize the penalty $-c$ and maximize the score, we find the threshold $k$ in test dataset value should be approximately 425 according to the empirical findings.

To distinguish unanswerable SQL statements, we assume that tokens of each generated SQL with low log probability are likely candidates for unanswerability, considering the log probabilities of the tokens as confidence scores. Since we previously determined the number of unanswerable candidates, or the threshold, we calculate the log probabilities of each SQL token in the test set items, sort them by ascending order of average value of its log probability, and consider all items with indices from the first up to the threshold as unanswerable. We incorporate some additional tricks, taking into account the characteristics of the SQL statement. The given text2SQL task is considered a highly structured sequence-to-sequence task due to the nature of SQL query syntax, which is very structured compared to the form of the input. The SQL statement inferred from the model can be broadly divided into two parts: reserved words of SQL syntax such as SELECT, AS, BETWEEN; and entities and attributes. We consider that the model is more likely to hallucinate when generating entities and attributes than when generating reserved words. Hence, when calculating the log probability for each test set item, we exclude reserved words(tokens) and compute it for the remaining tokens. The excluded reserved words can be found in Appendix \ref{appd:reserved}. Moreover, to make the distinction between answerable and unanswerable even clearer based on log probability, we also impose a limitation on the value of lowest $t$ tokens($t=10$ in this case), guiding the calculation towards the average value of these lowest log probability tokens. The algorithm for calculate log probability with one individual data can be found in Algorithm \ref{alg:pth}.

\subsection{Grammatical Errors Filtering}
\label{sec:methods:ehf}
In the last stage, we execute generated answerable SQL queries filtered by ProbGate through given database, if there is an error when executing SQL queries, we consider them unanswerable. The necessity of this stage arises because grammatical errors that are not fully caught by the previous ProbGate stage can only be detected by actual execution Although the query might actually have an answer and could be an answerable example, we consider it unanswerable to avoid penalties. This is because we can convert the penalty for incorrect answers, the $-c$ score, into 0. Reflecting on real-world scenarios, generating a response from the model indicating it does not know the answer could be more beneficial for the model's robustness and safety than returning incorrect results.

\section{Results and Analysis}

\begin{figure*}[t]
  \
  \includegraphics[width=0.5\linewidth]{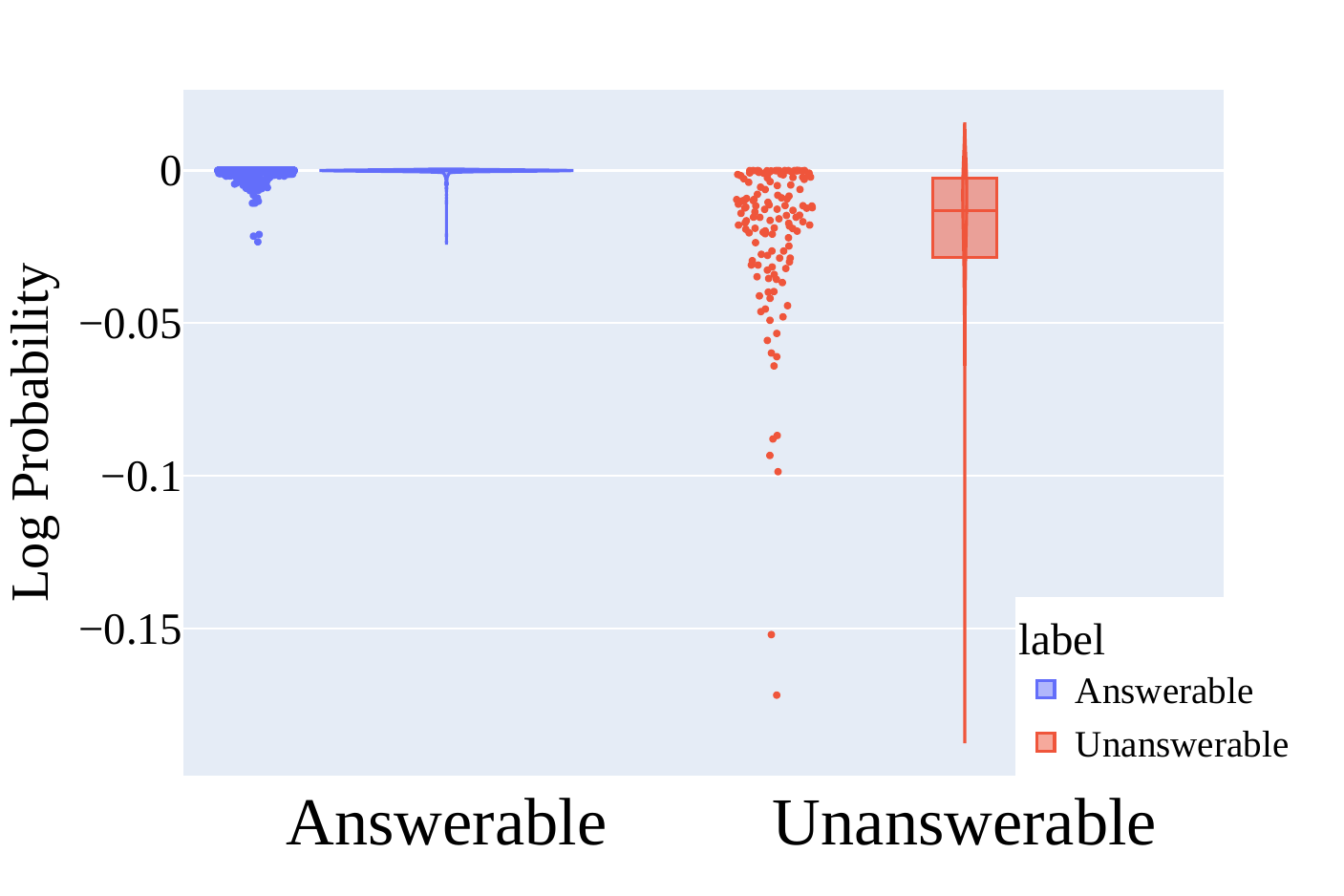} \hfill
  \includegraphics[width=0.5\linewidth]{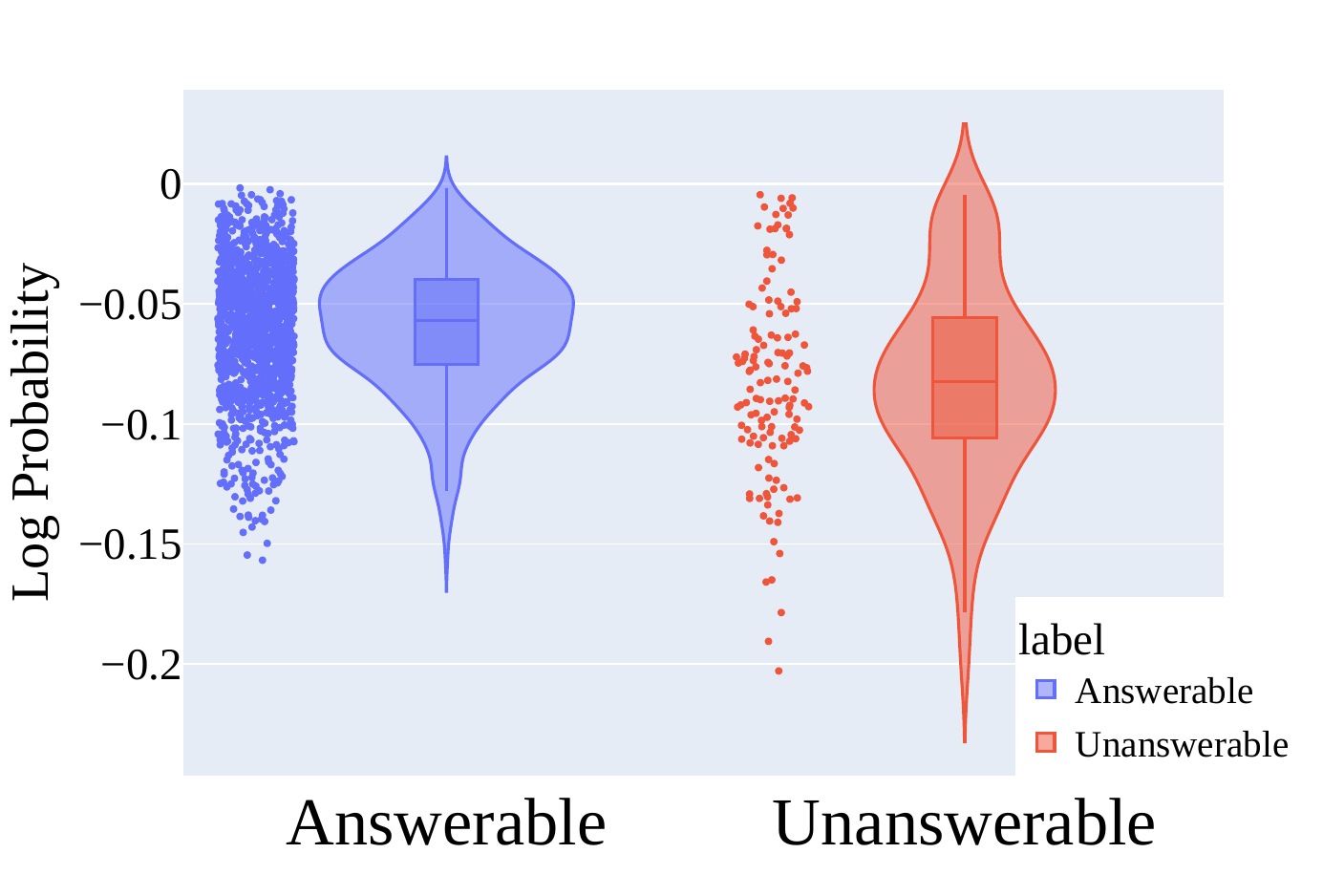}
  \caption {\textbf{Left} - Log Probability Distribution of the Fine-Tuned Model, \textbf{Right} - Log Probability Distribution of the Unfine-Tuned Model}
  \label{fig:distribution}
\end{figure*}

\subsection{Model Selection and Ablation Study}
\label{sec:methods:model}
As our final methodology, the base model gpt-3.5-turbo is relatively difficult to access the weights or perform additional analysis compared to other open-source models, so we use one of the Seq2Seq models, the T5 model, as a comparison model. Additionally, we compare using filtering based on maximum entropy, as utilized in \cite{lee2022ehrsql}, as our filtering model. Lastly, we also train a binary classifier with both T5 and gpt-3.5-turbo to distinguish between answerable and unanswerable questions to see its impact on performance. 
The results are shown in Table \ref{tab:ablation}, and conclusively, none of the methodologies surpasses the performance of the methodology applying gpt-3.5-turbo FT + ProbGate. The reason for this is observed in the accuracy of Text2SQL, where the gpt-3.5-turbo model, with its larger parameters and more advanced tuning methods, outperforms models from the T5 series. Additionally, it is interpreted that the Classifier does not show significant effectiveness due to the too different distribution between the training and the remaining dataset, and the task's high penalty for errors.

\subsection{ProbGate and Grammatical Errors Filtering}

The best results for the test set are achieved using our pipeline architecture, as shown in Table \ref{tab:final_score}. The process involves fine-tuning the gpt-3.5-turbo model with data excluding unanswerable data, then prioritizing the filtering of unanswerable data with ProbGate set to a threshold of 425, and finally applying Grammatical Errors Filtering. This sequence shows progressively better metric values. Additionally, we can interpret that the smaller the gap between the scores of RS\{0, 5, 10, N\}, the fewer penalties our model receives. Our final architecture can be seen as achieving the narrowest gap among these scores.

\subsection{Log Probability Distribution between Answerable and Unanswerable.}
In this section, we analyze the log probability distribution of SQL queries generated by the  gpt-3.5-turbo model and compare the differences in distribution based on whether the model is finetuned or not. For the experiments with the finetuned model, we first divide the training dataset into a 7:3 ratio, using 70\% of dataset to finetune  gpt-3.5-turbo with only answerable data. The remaining 30\% includes both answerable and unanswerable data, enabling the extraction of log probabilities during the model's SQL inference process. In left graph of Figure \ref{fig:distribution}, red represents null data, while blue indicates answerable data. The X-axis represents the log probability, and the Y-axis represents the number of data points with that log probability. As a result, it is observed that answerable data exhibited higher log probabilities, whereas null data show relatively lower probabilities, revealing the uncertainty in the generated SQL. The right graph of Figure \ref{fig:distribution} displays the log probability distribution of SQL generated by an unfine-tuned  gpt-3.5-turbo model under the same conditions. The difference in log probability distributions based on answerability is not significant, making it difficult to distinguish labels in the distribution. These results underscore the effectiveness of fine-tuning on answerable data, indicating that fine-tuning significantly increases the log probability of the model for answerable data while also creating a discernible distribution difference with unanswerable data. By leveraging this distributional difference, ProbGate  suggests that by setting an optimal threshold to treat all data that is either unanswerable or has uncertain generation outcomes as unanswerable, it can enhance response stability and reliability.

\section{Conclusion}
We participate in the EHRSQL Shared Task on Reliable Text-to-SQL Modeling On Electronic Health Records, as detailed in \cite{lee2024overview}, aiming to develop a reliable and high-performance Text2SQL method. This encompasses the challenge of generating appropriate SQL for answerable questions while also distinguishing unanswerable questions within datasets that include them. To solve this, we fine-tune LLMs on the training dataset and then employ a filtering pipeline called ProbGate, which consists of a combination of probabilistic threshold filtering and grammatical errors filtering, effectively executing the task. Additionally, through an ablation study and detailed analysis, we demonstrate that our method can be effectively used for tasks with a high sensitivity to errors. Ultimately, using this method, we conclude the shared task with a team ranking of 3rd place.

\section*{Limitations}
The methodology discussed here is central to solving competitive, contest-style shared tasks, with discussions taking place at a time when labels for the development and test sets, excluding training data, have not been disclosed. Therefore, our methodology greedily constructs the architecture to maximize the score on the main evaluation metric of the shared task, RS(10). Consequently, the primary parameters used in the model (e.g., threshold value, t value of ProbGate, etc.) can be specifically adjusted for the data and are sensitive to new datasets, meaning parameter values have a significant impact on the overall performance of the architecture. The performance of the basic model, which depends on the performance of the Fine-tuning model, is tied to a specific model (gpt-3.5-turbo) that is not open-sourced. Therefore, additional experiments with Text2SQL specialized open-source LLMs\cite{li2023starcoder} are needed.
These limitations increase in severity when the distribution of unanswered questions differs between training and test datasets. Therefore, further research on unanswered question filtering approaches from an out-of-distribution detection perspective is warranted.

\section*{Ethics Statement}
Throughout this research, we are using the gpt-3.5-turbo model as a baseline. It's acknowledged that depending on the inputs provided by users, the model's outputs may include harmful content or exhibit unintended biases. Recognizing and addressing these potential issues is essential for deploying this technology in real-world production environments. This entails a necessity for additional engineering tuning aimed at minimizing such side effects, highlighting a commitment to responsible AI use and the importance of continual improvement to ensure ethical deployment. Furthermore, the gpt-3.5-turbo model, which is used as our primary method, has not publicly disclosed its weights or training processes. There is also a risk that private data may be exposed during finetuning. Therefore, when handling sensitive data, it is advisable to switch to an open-source model or exercise caution.

\section*{Acknowledgments}
This work was supported by Artificial intelligence industrial convergence cluster development project funded by the Ministry of Science and ICT(MSIT, Korea)\&Gwangju Metropolitan City.

\bibliography{0.main}

\begin{thebibliography}{23}
\providecommand{\natexlab}[1]{#1}

\bibitem[{Brown et~al.(2020)Brown, Mann, Ryder, Subbiah, Kaplan, Dhariwal, Neelakantan, Shyam, Sastry, Askell et~al.}]{brown2020language}
Tom Brown, Benjamin Mann, Nick Ryder, Melanie Subbiah, Jared~D Kaplan, Prafulla Dhariwal, Arvind Neelakantan, Pranav Shyam, Girish Sastry, Amanda Askell, et~al. 2020.
\newblock Language models are few-shot learners.
\newblock \emph{Advances in neural information processing systems}, 33:1877--1901.

\bibitem[{Cao et~al.(2023)Cao, Chen, Li, Zhang, Xu, Zhang, and Yu}]{syntax_text2sql}
Ruisheng Cao, Lu~Chen, Jieyu Li, Hanchong Zhang, Hongshen Xu, Wangyou Zhang, and Kai Yu. 2023.
\newblock \href {https://doi.org/10.1109/TPAMI.2023.3298895} {A heterogeneous graph to abstract syntax tree framework for text-to-sql}.
\newblock \emph{IEEE Transactions on Pattern Analysis and Machine Intelligence}, 45(11):13796--13813.

\bibitem[{Deng et~al.(2022)Deng, Chen, and Zhang}]{deng-etal-2022-recent}
Naihao Deng, Yulong Chen, and Yue Zhang. 2022.
\newblock \href {https://aclanthology.org/2022.coling-1.190} {Recent advances in text-to-{SQL}: A survey of what we have and what we expect}.
\newblock In \emph{Proceedings of the 29th International Conference on Computational Linguistics}, pages 2166--2187, Gyeongju, Republic of Korea. International Committee on Computational Linguistics.

\bibitem[{Jiang et~al.(2023)Jiang, Xu, Gao, Sun, Liu, Dwivedi-Yu, Yang, Callan, and Neubig}]{jiang-etal-2023-active}
Zhengbao Jiang, Frank Xu, Luyu Gao, Zhiqing Sun, Qian Liu, Jane Dwivedi-Yu, Yiming Yang, Jamie Callan, and Graham Neubig. 2023.
\newblock \href {https://doi.org/10.18653/v1/2023.emnlp-main.495} {Active retrieval augmented generation}.
\newblock In \emph{Proceedings of the 2023 Conference on Empirical Methods in Natural Language Processing}, pages 7969--7992, Singapore. Association for Computational Linguistics.

\bibitem[{Kadavath et~al.(2022)Kadavath, Conerly, Askell, Henighan, Drain, Perez, Schiefer, Hatfield-Dodds, DasSarma, Tran-Johnson et~al.}]{kadavath2022language}
Saurav Kadavath, Tom Conerly, Amanda Askell, Tom Henighan, Dawn Drain, Ethan Perez, Nicholas Schiefer, Zac Hatfield-Dodds, Nova DasSarma, Eli Tran-Johnson, et~al. 2022.
\newblock Language models (mostly) know what they know.
\newblock \emph{arXiv preprint arXiv:2207.05221}.

\bibitem[{Katsogiannis-Meimarakis and Koutrika(2023)}]{text2sql_survey_2}
George Katsogiannis-Meimarakis and Georgia Koutrika. 2023.
\newblock \href {https://doi.org/10.1007/s00778-022-00776-8} {A survey on deep learning approaches for text-to-sql}.
\newblock \emph{The VLDB Journal}, 32(4):905--936.

\bibitem[{Lee et~al.(2024{\natexlab{a}})Lee, Chay, Cho, and Choi}]{lee2024trustsql}
Gyubok Lee, Woosog Chay, Seonhee Cho, and Edward Choi. 2024{\natexlab{a}}.
\newblock Trustsql: A reliability benchmark for text-to-sql models with diverse unanswerable questions.
\newblock \emph{arXiv preprint arXiv:2403.15879}.

\bibitem[{Lee et~al.(2022)Lee, Hwang, Bae, Kwon, Shin, Yang, Seo, Kim, and Choi}]{lee2022ehrsql}
Gyubok Lee, Hyeonji Hwang, Seongsu Bae, Yeonsu Kwon, Woncheol Shin, Seongjun Yang, Minjoon Seo, Jong-Yeup Kim, and Edward Choi. 2022.
\newblock Ehrsql: A practical text-to-sql benchmark for electronic health records.
\newblock \emph{Advances in Neural Information Processing Systems}, 35:15589--15601.

\bibitem[{Lee et~al.(2024{\natexlab{b}})Lee, Kweon, Bae, and Choi}]{lee2024overview}
Gyubok Lee, Sunjun Kweon, Seongsu Bae, and Edward Choi. 2024{\natexlab{b}}.
\newblock Overview of the ehrsql 2024 shared task on reliable text-to-sql modeling on electronic health records.
\newblock In \emph{Proceedings of the 6th Clinical Natural Language Processing Workshop}, Mexico City, Mexico. Association for Computational Linguistics.

\bibitem[{Li et~al.(2023)Li, allal, Zi, Muennighoff, Kocetkov, Mou, Marone, Akiki, LI, Chim, Liu, Zheltonozhskii, Zhuo, Wang, Dehaene, Lamy-Poirier, Monteiro, Gontier, Yee, Umapathi, Zhu, Lipkin, Oblokulov, Wang, Murthy, Stillerman, Patel, Abulkhanov, Zocca, Dey, Zhang, Bhattacharyya, Yu, Luccioni, Villegas, Zhdanov, Lee, Timor, Ding, Schlesinger, Schoelkopf, Ebert, Dao, Mishra, Gu, Anderson, Dolan-Gavitt, Contractor, Reddy, Fried, Bahdanau, Jernite, Ferrandis, Hughes, Wolf, Guha, Werra, and de~Vries}]{li2023starcoder}
Raymond Li, Loubna~Ben allal, Yangtian Zi, Niklas Muennighoff, Denis Kocetkov, Chenghao Mou, Marc Marone, Christopher Akiki, Jia LI, Jenny Chim, Qian Liu, Evgenii Zheltonozhskii, Terry~Yue Zhuo, Thomas Wang, Olivier Dehaene, Joel Lamy-Poirier, Joao Monteiro, Nicolas Gontier, Ming-Ho Yee, Logesh~Kumar Umapathi, Jian Zhu, Ben Lipkin, Muhtasham Oblokulov, Zhiruo Wang, Rudra Murthy, Jason~T Stillerman, Siva~Sankalp Patel, Dmitry Abulkhanov, Marco Zocca, Manan Dey, Zhihan Zhang, Urvashi Bhattacharyya, Wenhao Yu, Sasha Luccioni, Paulo Villegas, Fedor Zhdanov, Tony Lee, Nadav Timor, Jennifer Ding, Claire~S Schlesinger, Hailey Schoelkopf, Jan Ebert, Tri Dao, Mayank Mishra, Alex Gu, Carolyn~Jane Anderson, Brendan Dolan-Gavitt, Danish Contractor, Siva Reddy, Daniel Fried, Dzmitry Bahdanau, Yacine Jernite, Carlos~Mu{\~n}oz Ferrandis, Sean Hughes, Thomas Wolf, Arjun Guha, Leandro~Von Werra, and Harm de~Vries. 2023.
\newblock \href {https://openreview.net/forum?id=KoFOg41haE} {Starcoder: may the source be with you!}
\newblock \emph{Transactions on Machine Learning Research}.
\newblock Reproducibility Certification.

\bibitem[{Li et~al.(2022)Li, Li, Shang, Dong, Sun, Liu, Ji, Jiang, and Liu}]{li-etal-2022-pre}
Shaobo Li, Xiaoguang Li, Lifeng Shang, Zhenhua Dong, Chengjie Sun, Bingquan Liu, Zhenzhou Ji, Xin Jiang, and Qun Liu. 2022.
\newblock \href {https://doi.org/10.18653/v1/2022.findings-acl.136} {How pre-trained language models capture factual knowledge? a causal-inspired analysis}.
\newblock In \emph{Findings of the Association for Computational Linguistics: ACL 2022}, pages 1720--1732, Dublin, Ireland. Association for Computational Linguistics.

\bibitem[{Mellah et~al.(2020)Mellah, Ettifouri, Bouchentouf, and Belkasmi}]{text2sql_survey_1}
Youssef Mellah, Hassane~El Ettifouri, Toumi Bouchentouf, and Mohammed~Ghaouth Belkasmi. 2020.
\newblock Artificial neural networks for text-to-sql task: State of the art.
\newblock In \emph{Advances in Smart Technologies Applications and Case Studies}, pages 557--565, Cham. Springer International Publishing.

\bibitem[{Min et~al.(2023)Min, Ross, Sulem, Veyseh, Nguyen, Sainz, Agirre, Heintz, and Roth}]{lm_survey}
Bonan Min, Hayley Ross, Elior Sulem, Amir Pouran~Ben Veyseh, Thien~Huu Nguyen, Oscar Sainz, Eneko Agirre, Ilana Heintz, and Dan Roth. 2023.
\newblock \href {https://doi.org/10.1145/3605943} {Recent advances in natural language processing via large pre-trained language models: A survey}.
\newblock \emph{ACM Comput. Surv.}, 56(2).

\bibitem[{Park et~al.(2021)Park, Cho, Lee, Choo, and Choi}]{kg_text2sql}
Junwoo Park, Youngwoo Cho, Haneol Lee, Jaegul Choo, and Edward Choi. 2021.
\newblock Knowledge graph-based question answering with electronic health records.
\newblock In \emph{Machine Learning for Healthcare Conference}, pages 36--53. PMLR.

\bibitem[{Pourreza and Rafiei(2024)}]{din_sql}
Mohammadreza Pourreza and Davood Rafiei. 2024.
\newblock Din-sql: Decomposed in-context learning of text-to-sql with self-correction.
\newblock \emph{Advances in Neural Information Processing Systems}, 36.

\bibitem[{Raffel et~al.(2020)Raffel, Shazeer, Roberts, Lee, Narang, Matena, Zhou, Li, and Liu}]{2020t5}
Colin Raffel, Noam Shazeer, Adam Roberts, Katherine Lee, Sharan Narang, Michael Matena, Yanqi Zhou, Wei Li, and Peter~J. Liu. 2020.
\newblock \href {http://jmlr.org/papers/v21/20-074.html} {Exploring the limits of transfer learning with a unified text-to-text transformer}.
\newblock \emph{Journal of Machine Learning Research}, 21(140):1--67.

\bibitem[{Roziere et~al.(2023)Roziere, Gehring, Gloeckle, Sootla, Gat, Tan, Adi, Liu, Remez, Rapin et~al.}]{roziere2023code}
Baptiste Roziere, Jonas Gehring, Fabian Gloeckle, Sten Sootla, Itai Gat, Xiaoqing~Ellen Tan, Yossi Adi, Jingyu Liu, Tal Remez, J{\'e}r{\'e}my Rapin, et~al. 2023.
\newblock Code llama: Open foundation models for code.
\newblock \emph{arXiv preprint arXiv:2308.12950}.

\bibitem[{Shi et~al.(2024)Shi, Xu, Zhuang, Yu, Zhang, Wu, Zhu, Ho, Yang, and Wang}]{ehr_agent}
Wenqi Shi, Ran Xu, Yuchen Zhuang, Yue Yu, Jieyu Zhang, Hang Wu, Yuanda Zhu, Joyce Ho, Carl Yang, and May~D Wang. 2024.
\newblock Ehragent: Code empowers large language models for complex tabular reasoning on electronic health records.
\newblock \emph{arXiv preprint arXiv:2401.07128}.

\bibitem[{Touvron et~al.(2023)Touvron, Martin, Stone, Albert, Almahairi, Babaei, Bashlykov, Batra, Bhargava, Bhosale et~al.}]{touvron2023llama}
Hugo Touvron, Louis Martin, Kevin Stone, Peter Albert, Amjad Almahairi, Yasmine Babaei, Nikolay Bashlykov, Soumya Batra, Prajjwal Bhargava, Shruti Bhosale, et~al. 2023.
\newblock Llama 2: Open foundation and fine-tuned chat models.
\newblock \emph{arXiv preprint arXiv:2307.09288}.

\bibitem[{Wang et~al.(2020)Wang, Shin, Liu, Polozov, and Richardson}]{rat-sql}
Bailin Wang, Richard Shin, Xiaodong Liu, Oleksandr Polozov, and Matthew Richardson. 2020.
\newblock {RAT-SQL}: Relation-aware schema encoding and linking for text-to-{SQL} parsers.
\newblock In \emph{Proceedings of the 58th Annual Meeting of the Association for Computational Linguistics}, pages 7567--7578, Online. Association for Computational Linguistics.

\bibitem[{Wang et~al.(2023)Wang, Gao, Li, and Lou}]{text2sql_unans}
Bing Wang, Yan Gao, Zhoujun Li, and Jian-Guang Lou. 2023.
\newblock \href {https://doi.org/10.18653/v1/2023.findings-acl.352} {Know what {I} don{'}t know: Handling ambiguous and unknown questions for text-to-{SQL}}.
\newblock In \emph{Findings of the Association for Computational Linguistics: ACL 2023}, pages 5701--5714, Toronto, Canada. Association for Computational Linguistics.

\bibitem[{Wang and Sennrich(2020)}]{wang-sennrich-2020-exposure}
Chaojun Wang and Rico Sennrich. 2020.
\newblock \href {https://doi.org/10.18653/v1/2020.acl-main.326} {On exposure bias, hallucination and domain shift in neural machine translation}.
\newblock In \emph{Proceedings of the 58th Annual Meeting of the Association for Computational Linguistics}, pages 3544--3552, Online. Association for Computational Linguistics.

\bibitem[{Xiao and Wang(2021)}]{xiao-wang-2021-hallucination}
Yijun Xiao and William~Yang Wang. 2021.
\newblock \href {https://doi.org/10.18653/v1/2021.eacl-main.236} {On hallucination and predictive uncertainty in conditional language generation}.
\newblock In \emph{Proceedings of the 16th Conference of the European Chapter of the Association for Computational Linguistics: Main Volume}, pages 2734--2744, Online. Association for Computational Linguistics.

\end{thebibliography}

\appendix
\onecolumn
\section{Reserved Words List}
\label{appd:reserved}

This refers to a list of reserved words in SQL that we used in our experiment. 
\begin{mdframed}[backgroundcolor=verylightgray]
\small
["SELECT", "AS", "IN", "COUNT", "FROM", "WHERE", "AND", "OR", "INSERT", "UPDATE", "DELETE", "CREATE", "DROP", "ALTER", "JOIN", "ON", "GROUP BY", "ORDER BY", "HAVING", "LIMIT", "UNION", "DISTINCT", "INDEX", "TABLE", "VIEW", "TRIGGER", "PRIMARY KEY", "FOREIGN KEY", "NULL", "NOT NULL", "UNIQUE", "CHECK", "DEFAULT", "INDEX", "SEQUENCE", "EXEC", "LIKE", "BETWEEN", "EXISTS", "CASE", "WHEN", "THEN", "ELSE", "END", "CAST", "CHAR", "VARCHAR", "BOOLEAN", "INTEGER", "DATE", "INTERVAL", "TIME", "TIMESTAMP", "YEAR", "MONTH", "DAY", "HOUR", "MINUTE", "SECOND", "ZONE", "CURRENT\_DATE", "CURRENT\_TIME", "CURRENT\_TIMESTAMP", "TRUE", "FALSE"]
\end{mdframed}

\section{Input and Output Format}
\label{appd:format}
This is the input and output format according to the training specifications of gpt-3.5 turbo.

\begin{lstlisting}[language=json]
{
    `messages': [
        {`role': `system', `content': `You are `SQLgpt', an AI designed to convert natural language questions into their corresponding SQL queries. Your primary goal is to accurately generate the exact SQL query for each question presented to you.'},

        {`role': `user', `content': <Answerable Question>}, 

        {`role': `assistant', `content': <Correct SQL>}
    ]
}
\end{lstlisting}
\end{document}